\documentclass[letterpaper, 10 pt, conference]{ieeeconf}  
\IEEEoverridecommandlockouts                 
\usepackage{listings}
\usepackage{amsmath}
\usepackage{algorithm}
\usepackage[noend]{algpseudocode}
\usepackage{textcomp,gensymb}
\usepackage{graphicx}
\usepackage{xcolor}
\usepackage{soul}
\usepackage[colorlinks=true, urlcolor=blue, linkcolor=red]{hyperref}

\title{\LARGE \bf Robust Ladder Climbing with a Quadrupedal Robot
} 

\begin{document}
\author
{Dylan Vogel$^{*}$, Robert Baines$^{*}$, Joseph Church, Julian Lotzer, Karl Werner, and Marco Hutter
\noindent  \thanks{$^*$ Equal contribution}
\noindent  \thanks{{All authors are with ETH Zurich, Robotics Systems Lab; Leonhardstrasse 21, 8092 Zurich, Switzerland.}
         Contact: {\tt\small dyvogel@ethz.ch}}%
}

\maketitle
\thispagestyle{empty}
\pagestyle{empty}

\newcommand{\mrm}[1]{\mathrm{#1}}

\begin{abstract}

Quadruped robots are proliferating in industrial environments where they carry sensor payloads and serve as autonomous inspection platforms. Despite the advantages of legged robots over their wheeled counterparts on rough and uneven terrain, they are still unable to reliably negotiate a ubiquitous feature of industrial infrastructure: ladders.
Inability to traverse ladders prevents quadrupeds from inspecting dangerous locations, puts humans in harm's way, and reduces industrial site productivity.
In this paper, we learn quadrupedal ladder climbing via a reinforcement learning-based control policy and a complementary hooked end effector. We evaluate the robustness in simulation across different ladder inclinations, rung geometries, and inter-rung spacings. On hardware, we demonstrate zero-shot transfer with an overall 90\% success rate at ladder angles ranging from 70$^{\circ}$ to 90$^{\circ}$, consistent climbing performance during unmodeled perturbations, and climbing speeds 232$\times$ faster than the state of the art. 
This work expands the scope of industrial quadruped robot applications beyond inspection on nominal terrains to challenging infrastructural features in the environment, highlighting synergies between robot morphology and control policy when performing complex skills. 
More information can be found at the project website: \href{https://sites.google.com/leggedrobotics.com/climbingladders}{https://sites.google.com/leggedrobotics.com/climbingladders.}
\end{abstract}


\section{INTRODUCTION}

Historically, most quadruped robots have been limited to locomotion on highly structured terrains~\cite{li_research_2011}. 
Advances in control algorithms and hardware over the past decade have resulted in highly agile systems capable of stable locomotion over irregular natural surfaces~\cite{miki_learning_2022,shafiee_viability_2024} and ``parkour'' over obstacles~\cite{park_jumping_2021, hoeller_parkour_2024}. 
Thanks to their ability to traverse rough terrain and remain stable on uneven surfaces, quadruped robots are becoming commonplace at industrial sites where they perform routine inspection tasks that are dangerous or undesirable for humans~\cite{fan_review_2024}. 
Despite ongoing advances, quadruped robots are still unable to robustly traverse many types of infrastructure common in man-made environments, a key example being ladders.

 \begin{figure}
 \centering
 \includegraphics[width = 0.48\textwidth]{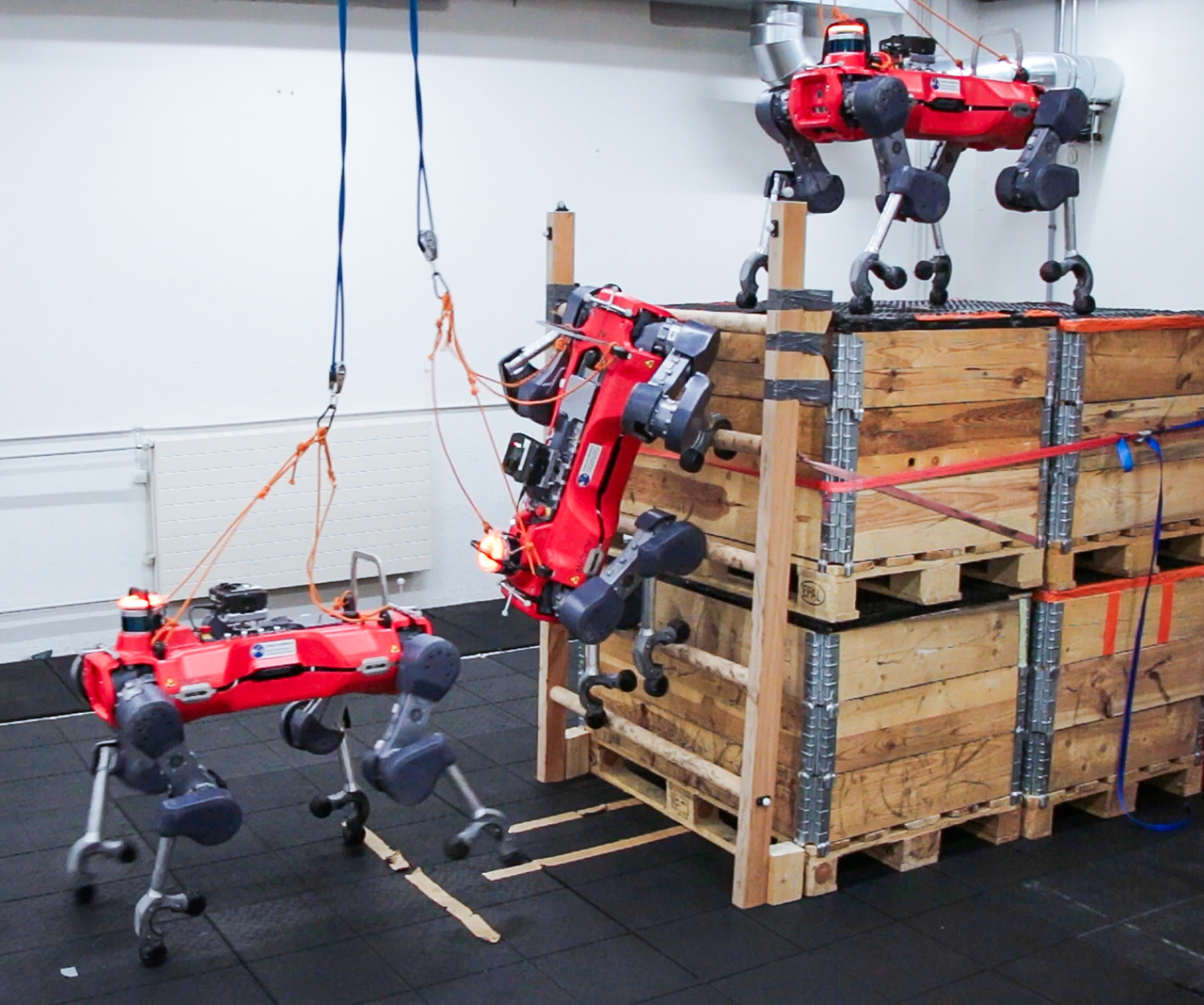}
 \caption{Composite image of a quadrupedal robot equipped with hooked end effectors, ascending a ladder in 4~s with a reinforcement learning-based control policy. Ladder shown has parameters 90$^{\circ}$ incline, 1.8~m length,  0.3~m inter-rung spacing, 2.5~cm rung radius, and 1~m width.}
 \label{fig:splash} 
 \end{figure}

Ladder falls are a major occupational hazard; the American Academy of Orthopedic Surgeons estimates that 500,000 ladder-related injuries occur per year in the United States alone~\cite{NationalLadderSafetyMonth2024}. 
To minimize occupational risk and improve site efficiency, next-generation robots must be able to robustly negotiate all types of industrial terrain, including ladders.
However, quadrupeds are not normally equipped with the appropriate morphology or control strategy for such a task. 
Consider that quadrupeds typically have ball-shaped or flat feet, preventing the anchoring forces needed for reliable upward movement~\cite{catalano_adaptive_2022}. Ladder climbing also requires full-body coordination to stabilize the center of mass and ascend steep inclines~\cite{hammer_human_1992}, a challenge for robots in unmodeled environments~\cite{sun_stable_2021}.

Previous research on robotic ladder climbing has been conducted mainly with humanoid robots, is characterized by very slow climbing speeds, and has not generalized beyond specific ladder configurations in structured environments (e.g., completely vertical ladders and no perturbations)~\cite{vaillant_multi-contact_2016}. 
In one such work, a humanoid equipped with dexterous hands and an \textit{a priori} motion trajectory climbed a vertical ladder~\cite{yoneda_vertical_2008}.
Extensions to this work validated the robot on several different vertical ladders, but did not demonstrate improved vertical speed beyond $\sim$26~mm/h or robustness to perturbations~\cite{yoshiike_development_2017}. 
In another work, a motion planner and compliance controller were combined to generate disturbance-resistant climbing trajectories. However, in the one ladder example demonstrated on hardware, the robot took seven minutes to traverse only five rungs~\cite{luo_robust_2014}.
Ladder climbing has also been demonstrated on a couple of quadrupeds~\cite{sun_planning_2017, saputra_novel_2019}, although only vertical ladders were considered and the robots' movements remained slow, taking up to two minutes to ascend a single rung. 

Outside of ladder climbing specifically, robust locomotion of quadrupeds has been demonstrated in other challenging environments. 
Model-based methods, built around non-linear model predictive control or other trajectory optimization methods, typically excel in sparse terrains such as stepping stones and gaps~\cite{grandia_perceptive_2023}. However, such methods are vulnerable to modeling uncertainties, external disturbances, and degraded perception. In contrast, model-free methods such as reinforcement learning (RL) have shown great simulation to reality (sim2real) transfer, real-world robustness over rough landscapes~\cite{miki_learning_2022, kim_not_2023}, and steady progress on sparse terrain problems such as stepping stones~\cite{jenelten_dtc_2023, agarwal_legged_2022} and parkour~\cite{hoeller_parkour_2024}.

In this paper, we propose a model-free RL control policy and a complementary hooked end effector that enables fast and generalizable quadrupedal ladder climbing (Fig.~\ref{fig:splash}; Supplementary video). 
We extend \cite{miki_learning_2022} and use a privileged teacher-student training setup to first learn a robust teacher policy with access to noiseless information about the environment and external disturbances. We then distill this behavior into a student policy which only has access to on-board observations, but is trained to reconstruct the privileged information via a recurrent memory architecture. 

\looseness=-1
Although elevation mapping is a common sensing modality for rough terrain locomotion \cite{miki_learning_2022, kim_not_2023, jenelten_dtc_2023}, it fails in the case of vertical ladders (which appear as a line when viewed from above). In this paper, we focus on locomotion, so in lieu of more complex perception methods such as depth cameras \cite{agarwal_legged_2022} or terrain reconstruction \cite{hoeller_parkour_2024}, we use a motion capture system to provide the ladder position to the control policy. 
In summary, the contributions of this paper include:
\begin{itemize} 
\item An extension of \cite{miki_learning_2022} to generate robust control policies for quadrupedal ladder climbing with model-free RL.
\item A hook end effector design that generates the necessary forces for reliable and robust climbing. 
\item Extensive simulations of successful climbing across diverse ladders parameterized by length $L_\mrm{len}$, width $L_\mrm{width}$, inter-rung spacing $L_\mrm{space}$, rung radius $L_\mrm{radius}$, and inclination angle $L_\mrm{\theta}$. 
\item Hardware demonstrations of the fastest and most generalized robotic ladder climbing to date, tested on ladders with varying $L_\mrm{\theta}$ and under unmodeled perturbations. 
\end{itemize}


\section{Hook design}

\begin{figure}
 \centering
 \includegraphics[width = 0.48\textwidth]{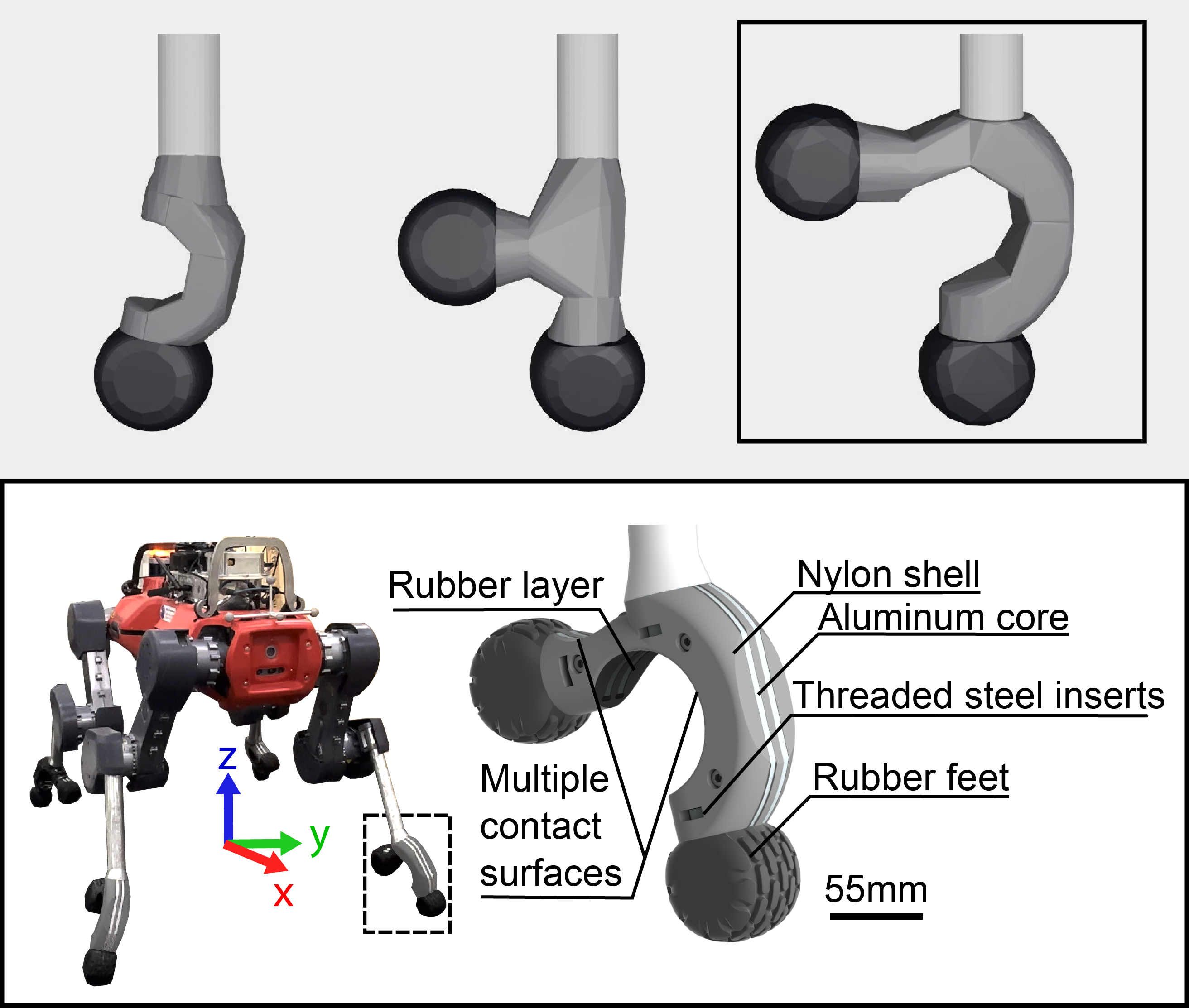}
 \caption{Top: Different hook designs screened in simulation. Bottom: Quadruped robot used for testing, equipped with the best-performing hooked end effector. Composed of aluminum cores and 3D printed shells, the hook features concave surfaces that encourage stable poses on round rungs, as well as protrusions that allow pushing and pulling on the rungs. The robot's base frame is defined with its origin at the center of the torso and the illustrated coordinate directions.}
 \label{fig:hook} 
 \end{figure}

A hooked end effector was developed to encourage the emergence of robust climbing behavior. Multiple design variations were heuristically evaluated and iteratively refined through simulation (Fig.~\ref{fig:hook}: Top). The final design incorporates strategically contoured concave regions, enabling the robot to exert both compressive and tensile forces on the rungs (Fig.~\ref{fig:hook}: Bottom). This configuration facilitates stable postures, even when the robot's center of mass extends beyond its support triangle. The hook's large concave surface allows passively stable ladder engagement over a large range of angles, a feature that proved critical for maintaining performance under external disturbances. 
\section{Control Methods}

\begin{figure*}
 \centering
 \includegraphics[width = 0.98\textwidth]{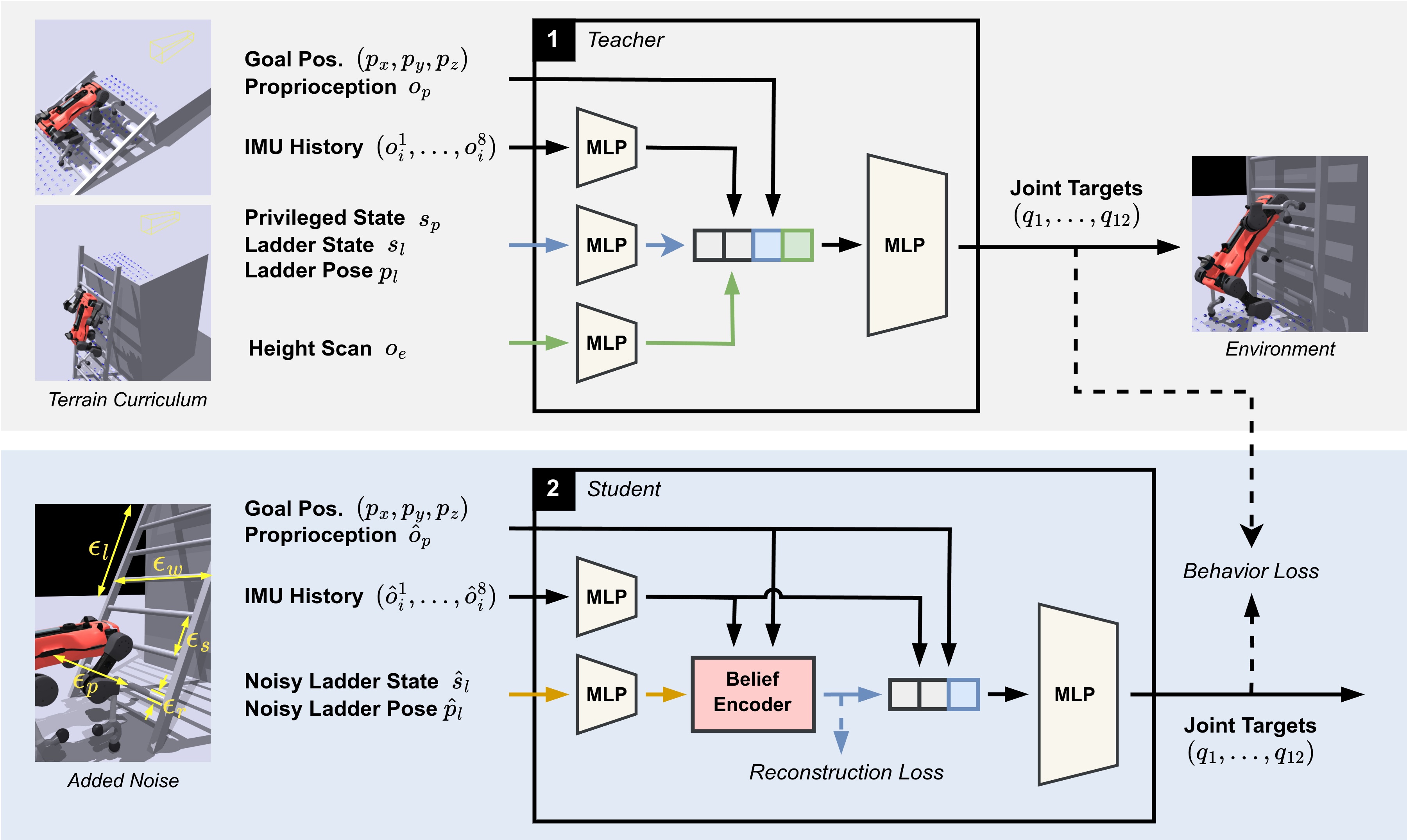}
 \caption{Illustration of the network architecture and training pipeline. First, a teacher policy is trained in simulation with access to noiseless observations and privileged state information. Then, a student policy with a recurrent belief encoder is trained to mimic the teacher actions and reconstruct the true privileged state from noisy observations and a noisy estimate of the ladder state and pose. The student policy is deployed on the actual robot hardware.}
 \label{fig:architecture} 
 \end{figure*}

The general training pipeline is illustrated in Fig.~\ref{fig:architecture} and follows that of reference~\cite{miki_learning_2022}. We first train a teacher policy in simulation with access to noiseless proprioceptive observations $o_p$, noiseless inertial measurement unit (IMU) history $o_i^H$, a height scan around the robot $o_e$, and privileged state information $s_p$. We then distill a student policy that only has access to noisy onboard proprioceptive measurements $\hat{o}_p$, IMU history $\hat{o}_i^H$, ladder state $\hat{s}_l$, and ladder pose $\hat{p}_l$. 
The resulting student policy outputs joint position targets $q_i$ at 50~Hz, which are tracked by a proportional-derivative controller running at 400~Hz on the real robot. A learned actuator network  models the joint dynamics in simulation~\cite{hwangbo_learning_2019}.

\textit{Policy Observations:}
Proprioceptive observations $o_p$ include the goal direction and heading in base frame, gravity direction, current joint positions, and a history of joint velocities and joint position tracking errors at 0~s, -0.02~s, and -0.04~s relative to the current timestep. A ``standing mode'' observation is provided which is true if the agent is within 15~cm of the goal, arbitrarily chosen. In lieu of base velocities, we directly provide a history of the last eight IMU measurements $(o_i^1, \dots, o_i^8)$ at 400~Hz, including base linear acceleration and angular velocity.

\looseness=-1
During teacher training, a height scan $o_e$ with dimensions 2x1~m and a resolution of 10~cm is provided around the robot, as it was found to accelerate policy learning. It is subsequently removed during student training.
Privileged state information $s_p$ includes body contact states, feet contact forces, friction coefficients, external forces and torques applied to the base and feet, mass added to the base, airtime of each foot, feet positions in base frame, a ladder state vector $s_l$, and ladder pose $p_l$. 
The ladder state comprises a binary flag indicating whether a ladder is present in the current terrain, $L_\mrm{\theta}$, $L_\mrm{width}$, $L_\mrm{radius}$, $L_\mrm{space}$, and number of rungs. The ladder pose consists of the position and yaw of the bottom rung in the robot base frame.
A noisy version of the ladder state $\hat{s}_l$ and pose $\hat{p}_l$ are given as exteroceptive observations during student training. During real-world deployment, the ladder state is measured directly, and the pose is estimated from motion capture data. Noise is added to the state in simulation to account for measurement error and irregularities in ladder geometry.

\looseness=-1
\textit{Teacher Policy Training:}
The teacher policy is similar to that of \cite{miki_learning_2022}, with the addition of a multi-layer perceptron that encodes the IMU observation history before concatenating it with proprioceptive observations. 
The policy is trained using Interior-point Policy Optimization (IPO) with adaptive constraint thresholding~\cite{kim_not_2023}. IPO is used to enforce constraints on the joint limits (position, velocity, torque), which was found to result in fewer violations compared to a reward penalty. Tab.~\ref{tab:reward_equations} \&~\ref{tab:symbols} summarizes the rewards. Episodes terminate if the robot base inclination exceeds 100$^{\circ}$ in pitch or roll. 

\looseness=-1
Two different types of terrain are randomly generated during training: i) rough terrain consisting of boxes and slopes, and ii) ladders of varying $L_\mrm{len}$, $L_\mrm{\theta}$, $L_\mrm{width}$, $L_\mrm{space}$, and $L_\mrm{radius}$. The training curriculum is adaptive, and agents progress to more difficult terrains (longer, steeper) as the agent reaches earlier goals \cite{pmlr-v164-rudin22a}. To simplify the curriculum, the rungs are parameterized as elliptic cylinders with a minor axis of 2.5~cm and a major axis that decreases with increasing curriculum difficulty, down to 2.5~cm. This value was chosen such that the rungs would be smaller than the 2.75~cm opening radius of the hook. Early-curriculum ladders therefore have broad, flat rungs that are easy to climb. The ladders are randomly offset from the end platform up to 15~cm, with a minimum clearance for the foot as $L_\theta$ increases.

Agents are spawned in a random configuration and commanded to reach a random goal position and heading. Unlike \cite{hoeller_parkour_2024}, we do not sample command times, as we find that i) it is difficult to estimate the amount of time required for climbing, and ii) a fixed command time prevents the agent from falling and learning robust retry behaviors. Instead, we use a dense tracking reward (see Tab.~\ref{tab:reward_equations}) and allow any time up to the episode length of 10~s. We find that this improves success rates compared to prior methods.
The agent is initialized using a state from the previous episode $50\%$ of the time, with randomization added to the base and joint velocities. We also apply random external forces and torques to the base of the robot, random velocity offsets (pushes), randomize the base mass, and randomize the friction coefficients of the feet.

\bgroup
\def\arraystretch{1.4}
\begin{table}
  \centering
  \caption{Reward Equations}
  \label{tab:reward_equations}
  \begin{tabular}{l|l}
    \hline
    \textbf{Name} & \textbf{Equation} \\ \hline
    Position Tracking & $3(\neg\delta_\mathrm{goal}(v_b \cdot \hat{p}_\mathrm{goal} - v_{b,\mathrm{over}}^2) + 1.5\delta_\mathrm{goal})$ \\
    Heading Tracking & $0.5\exp(-10(\psi_\mathrm{goal} - \psi_b)^2)\exp(-4||p_\mathrm{goal}||_2^2)$ \\
    Base Motion & $0.2 (\exp(-v_{b,z}^2) + \exp(-0.5(\dot{\phi}_b^2 + \dot{\theta}_b^2))) $ \\
    Joints & $-0.001\sum_{i=1}^{12} (0.01 \tau_i^2 + \dot{q}_i^2 + 0.2\ddot{q}_i)$\\
    Action Rate & $-0.01\sum_{i=1}^{12} (q_{i,t}^* - q_{i,t-1}^*)^2$ \\
    Action Smoothness & $-0.01 \sum_{i=1}^{12} (q_{i,t-2}^* - 2q_{i,t-1}^* + q_{i,t}^*)^2$ \\
    Foot Slippage & $-0.25\sum_{k\in\mathrm{feet}} c_k||v_k||_2(1 - 0.8 \mathbf{1}(\mu_k < 0.5))$ \\
    Flat Orientation & $-\delta_f(\hat{g}_{b,x}^2+\hat{g}_{b,y}^2)(1 + 8\delta_\mathrm{goal})$ \\
    Stand Still & $-0.5\delta_f\delta_\mathrm{goal}\sum_{i=1}^{12} |q_i^* - q_{i,0}|$ \\
    Stand Still Contact & $-0.5\delta_\mathrm{goal}\sum_{k\in\mathrm{feet}} \neg c_k$ \\
    Collision & $-0.1\sum_{k\in\mathrm{thighs,shanks}} c_k$ \\
    Base Collision & $-c_\mathrm{base}$ \\
    \hline
  \end{tabular}
\end{table}
\egroup

\bgroup
\def\arraystretch{1.2}
\begin{table}
    \centering
    \caption{Symbols for Tab.~\ref{tab:reward_equations}}
    \label{tab:symbols}
    \begin{tabular}{l|l}
        \hline
        \textbf{Symbol} & \textbf{Description} \\ \hline
        $v_b$ & Velocity of the base in base frame \\
        $v_{b,\mathrm{over}}$ & $\max(0, ||v_b||_2 - 0.7)$\\
        $p_\mathrm{goal}, \hat{p}_\mathrm{goal}$ & Vector and unit vector from base to goal in base frame\\
        $\delta_\mathrm{goal}$ & $\mathbf{1}(||p_\mathrm{goal}||_2 < 0.15) $\\
        $\delta_f$ & 1 if the local terrain is flat, 0 otherwise \\
        $\psi_\mathrm{goal}$ & Yaw of the goal \\
        $\phi_b, \theta_b, \psi_b$ & Roll, pitch, and yaw of the base \\
        $q_i, q_i^*, q_{i,0}$ & Actual, desired, and default position of joint $i$ \\
        $c_k$ & 1 if body $k$ is in contact, 0 otherwise \\
        $v_k$ & Velocity of body $k$ in base frame \\
        $\mu_k$ & Friction coefficient of body $k$ \\
        $\hat{g}_b$ & Gravity direction in base frame \\
        $\mathbf{1}(\cdot)$ & Binary indicator function \\
        \hline
    \end{tabular}
\end{table}
\egroup

\textit{Student Policy Training:}
A student policy is trained using supervised learning to output the same actions as the teacher policy, while only having access to noisy versions of the observations.
A recurrent belief encoder is used to reconstruct the privileged observations provided to the teacher.
In this work, the noisy ladder state and pose are treated as exteroceptive observations that the student can reconstruct from contacts with the environment.

\textit{Simulation Setup:}
All simulations are performed in LeggedGym~\cite{pmlr-v164-rudin22a}. We train using 4096 parallel environments with 48 and 120 steps per batch for the teacher and student, respectively. The teacher is trained for 15,000 epochs and the student is trained for 5,000, taking around 4.5 days in total on a NVIDIA RTX 3090. Training with the hook end effector is around 30\% slower than with the ball foot due to the additional collision bodies, which are approximated by convex decomposition.

\section{Results}

 \begin{figure}
 \centering
 \includegraphics[width = 0.40\textwidth]{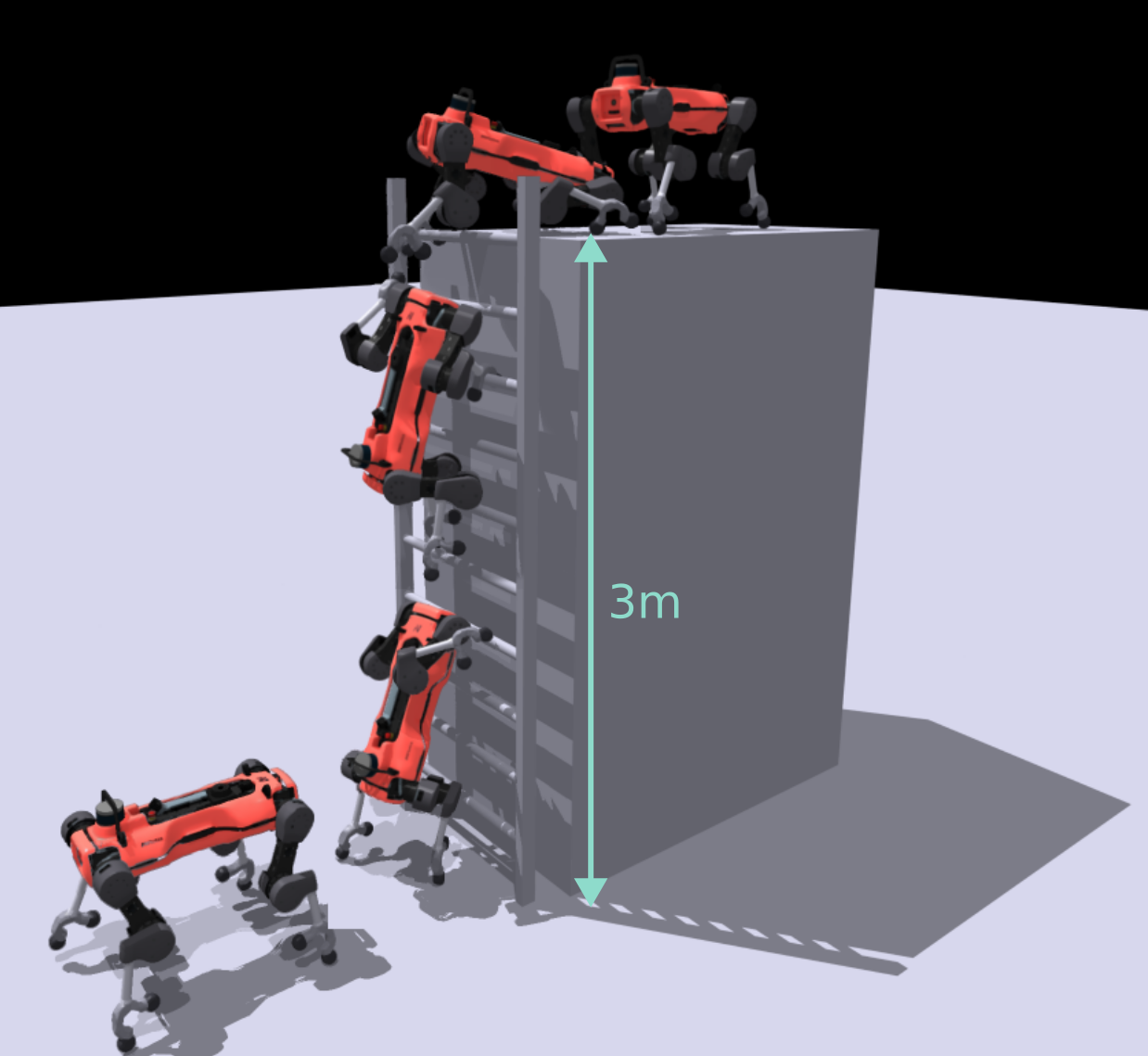}
 \caption{A simulated robot was assessed for its ability to traverse ladders of different configurations. Shown is a composite image generated in simulation of the longest ladder tested:  $ L_\mrm{len} =   \textnormal{3~m}$ at $ L_\mrm{\theta} = \textnormal{90}^{\circ}$.}
 \label{fig:sim_long_ladder} 
 \end{figure}

\textit{Simulation Results:}
The student policy was evaluated in simulation across various $L_\mrm{\theta}$ and $L_\mrm{radius}$. The policy was evaluated in the presence of observation noise and external disturbances, to provide a more realistic evaluation of policy robustness.
For evaluation, $L_\mrm{len}$ was randomly sampled between 1-3~m, $L_\mrm{width}$ between 1.0-1.25~m, and $L_\mrm{space}$ between 27.5-32.5~cm across 50 different ladders. Agents who reached the goal without terminating in under 15~s were marked as successful and the results were averaged among 3072 agents. The rungs were made purely cylindrical during the evaluation, rather than the elliptic cylinders used during training. An example evaluation is shown in Fig.~\ref{fig:sim_long_ladder}. 

\begin{figure}
\centering
\includegraphics[width = 0.48\textwidth]{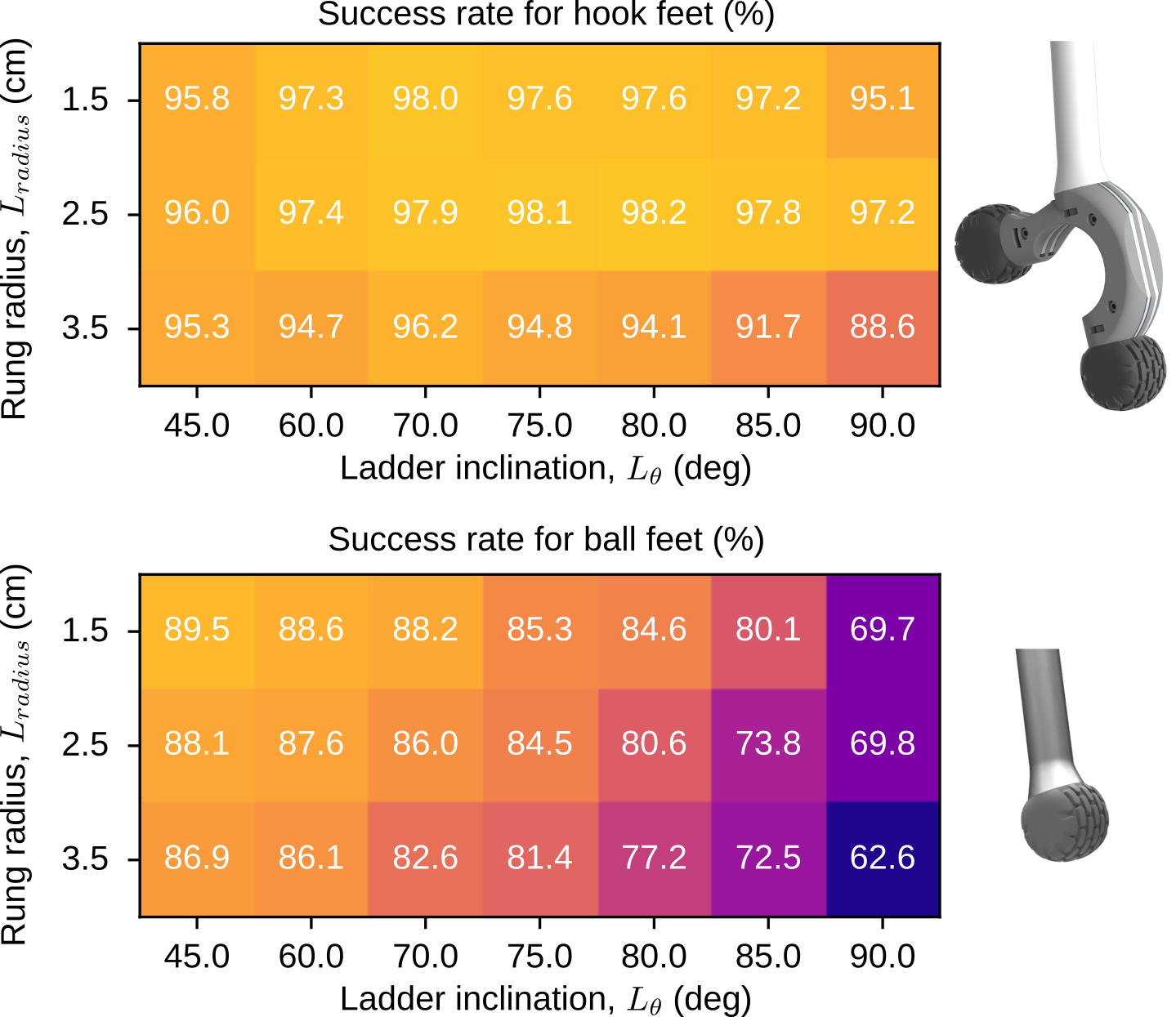}
\caption{Climbing success rate in simulation across various ladder inclinations and rung radii, juxtaposing performance of the ball and hooked end effector. The agents were evaluated in the presence of noise and external disturbances.}
\label{fig:sim_result_hook} 
\end{figure}

\looseness=-1
The robot with the hooked end effector achieved an average success rate 96\% across all configurations tested (see Fig.~\ref{fig:sim_result_hook}). 
The robot successfully climbed ladders in the presence of added observation noise and external disturbances, such as random velocity offsets applied to the base every 5~s, sampled from a normal distribution with a standard deviation of 1~m/s. 
In the case of $L_\mrm{radius} = \mathrm{3.5}~\textnormal{cm}$, which is larger than the opening radius of the hook, we observed a steady drop in performance at steeper $L_\mrm{\theta}$ as the agent could no longer stabilize itself with the front legs. As $L_\mrm{\theta}$ increases, the robot's applied forces switch from mostly compressive to a mix of compressive and tensile forces needed to stabilize the center of mass.

We compared the performance of the hook to a policy trained with the traditional ball-foot, observing a clear performance gap across all configurations and a notable performance drop at steeper $L_\mrm{\theta}$. The average success rate across all configurations was only 81\% for the ball-foot (Fig.~\ref{fig:sim_result_hook}).
Interestingly, the performance of the ball-foot design improved with decreasing $L_\mrm{radius}$ across all configurations. We observed that the policy can exploit the small crease between the foot and the shank to weakly anchor the robot on the rungs at smaller $L_\mrm{radius}$. Moreover, with decreasing $L_\mrm{radius}$, it became easier for the robot to maneuver its knees between the rungs, a movement strategy the policy relies on heavily in the absence of the hooked end effector.

To better understand the benefits of the hooked end effector, we ran noise-free evaluations over the configurations in Fig.~\ref{fig:sim_result_hook}. In this case, both designs yielded success rates greater than 99\% in all configurations. From this, we conclude that the hook furnishes stability to generalize over uncertainty and enables smooth and rapid climbing behavior---critical factors for robust real-world deployment.

\looseness=-1
\textit{Real World Results:}
The control policy was deployed zero-shot on an ANYmal D robot (ANYbotics AG) without further fine-tuning. A motion capture system was used to estimate the ladder pose, $p_l$, along with $L_\mrm{\theta}$. Other elements of the ladder state, $s_l$, were measured directly and input as scalars to the policy. 
In our setup, the reference ladder had parameters
$L_\mrm{width}= \mathrm{1}~\textnormal{m}$, 
$L_\mrm{len} =  \mathrm{1.8}~\textnormal{m} $, 
$L_\mrm{space} = \mathrm{30}~\textnormal{cm}$ (with five cylindrical rungs), 
and $L_\mrm{radius} = \mathrm{2.5}~\textnormal{cm}$. 
The ladder was placed at various $L_\mrm{\theta}$ against wooden boxes. 
At $L_\mrm{\theta}$ values of 70$^{\circ}$ and 80$^{\circ}$, the policy was successful in four of four tests in each configuration. At $L_\mrm{\theta} = \textnormal{90}^{\circ}$, the policy was successful in two of the three tests.
During the third test, unmodeled standoffs on the underside of the robot were observed to collide with the top rung. We decided not to proceed with further tests at 90$^{\circ}$ due to this obvious sim2real gap.

 \begin{figure}
 \centering
 \includegraphics[width = 0.48\textwidth]{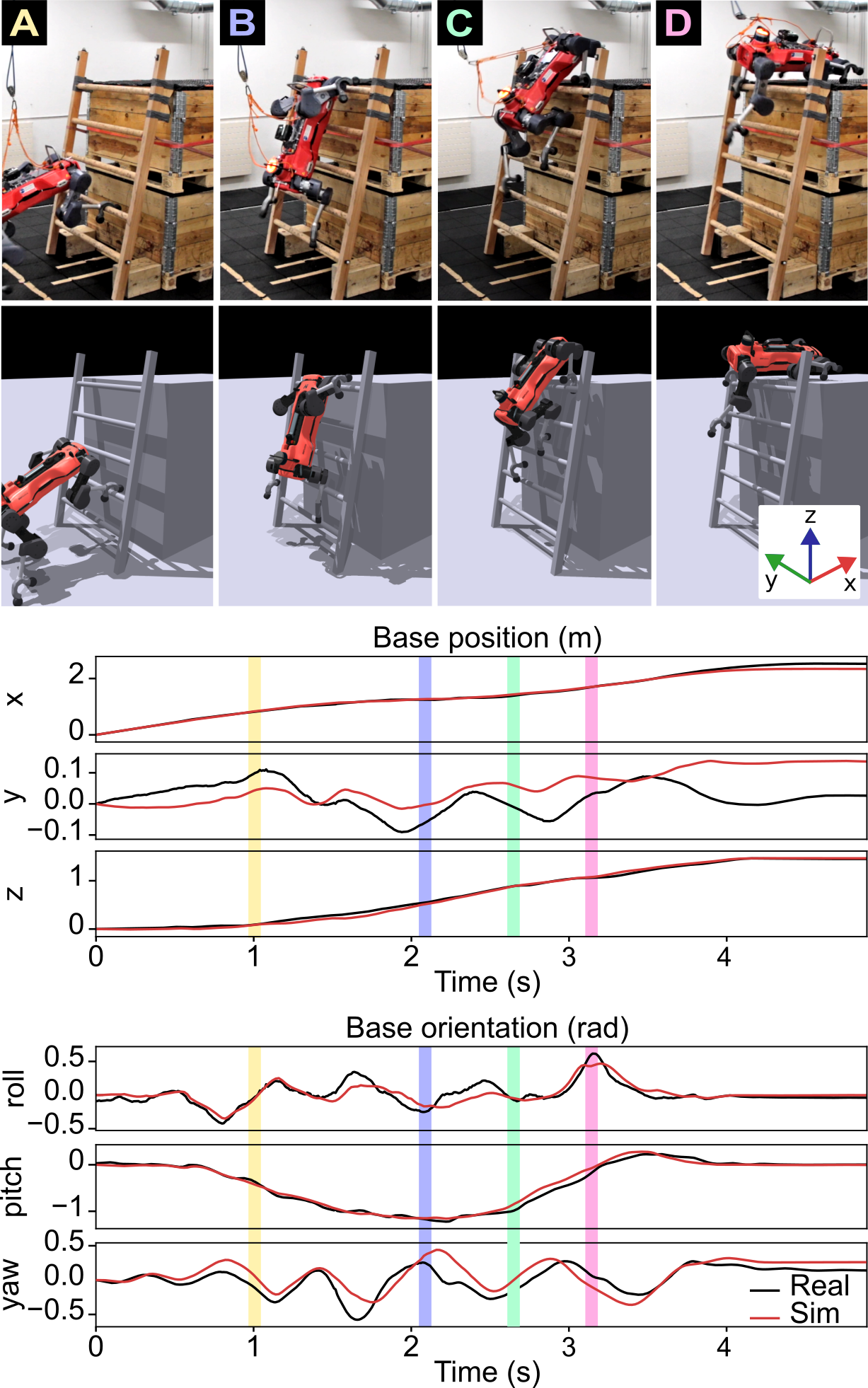}
 \caption{Representative sim2real experiment. Key snapshots highlighted: A) ladder mounting, B) midway point, C) approaching the dismount, and D) dismount. Plots show robot base position and orientation data over time while ascending a ladder with $L_\mrm{\theta} = \textnormal{80}^{\circ}$. Consistency between simulation and reality is furthermore observed throughout the experiment through low deviation between Real and Sim curves. }
 \label{fig:sim2real} 
 \end{figure}

A representative sim2real experiment with a ladder at $L_\mrm{\theta} = \mathrm{80^{\circ}}$ is shown in Fig.~\ref{fig:sim2real}.
Snapshots of key moments during climbing highlight the close similarity between simulation and real experiments. Joint positions and environmental contacts during mounting, passing the midway point, and dismounting---culminating in a rapid roll and flick of the back right leg---closely mirror the observed simulation. 
In this particular example, the position and orientation of the robot base had an average root mean squared error of 11~cm and 0.18~rad, respectively, over the entire trajectory.

The robot's average climbing speed was determined from video footage of the real experiments, using the time elapsed from first contact to the moment when the robot fully dismounted from the ladder. Over ten successful runs, the average speed was found to be 0.44~m/s (0.51~bl/s) with a standard deviation of 0.16~m/s. For context, our method yields 232$\times$ faster ladder climbing speeds than the existing state-of-the-art quadruped robot in reference~\cite{saputra_novel_2019}.

\begin{figure}
 \centering
 \includegraphics[width = 0.48\textwidth]{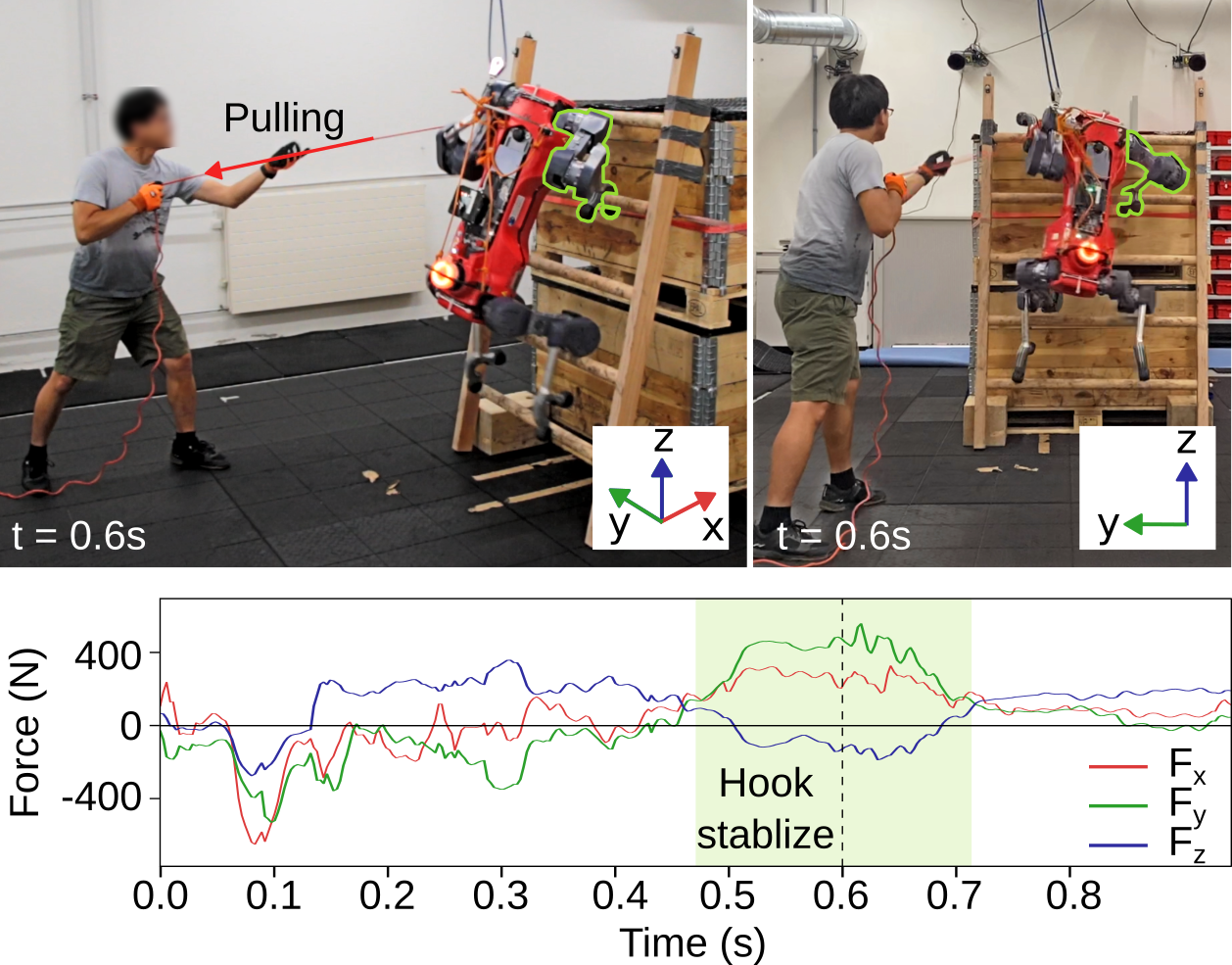}
 \caption{The robot successfully climbs ladders even in the presence of unmodeled perturbations. Shown is an excerpt from a test at 80$^{\circ}$ with a rope tied to the robot's front left foot. We pulled the rope at various points during the ascent. Due to the shape of the hook, the robot could successfully anchor itself, generating tension forces between its body and the rungs with its front right foot (seen as increase of $F_{x}$ and $F_{y}$ to the positive region of the plot and drop of $F_{z}$ to the negative region).}
 \label{fig:perturb} 
 \end{figure}

The policy was also tested for robustness to unmodeled perturbations. A rope was tied to different parts of the robot's chassis and feet, and pulled at different times during ascent. 
Fig.~\ref{fig:perturb} shows an experiment in which the rope was tied to the robot's left front foot. 
Analysis of the reaction forces over a characteristic window of time during which the robot was being pulled reveals that the robot can switch from pushing down on the rung (positive $F_z$) to generating tensile forces with the hook that it uses to brace itself. In other trials, the policy was found to exhibit persistent recovery and retry behaviors while being pulled from points on the base (Supplementary Video).

\section{Conclusion}

\looseness=-1
We demonstrated quadrupedal ladder climbing via a new robot end effector design and a complementary RL-derived control policy. 
The end effector's design allows for seamless walking on both nominal and rough terrain, while the controller naturally transitions between walking and climbing without requiring separate locomotion policies. Both the gait patterns and climbing behavior emerge entirely from the learned policy, effectively exploiting the custom hook-shaped end effector.

Performance assessments in simulation revealed an overall 96\% success rate at climbing ladders, even with disturbances.
Subsequent sim2real experiments validated that the proposed approach elicits robust and reliable policies on robot hardware. 
Crucially, compared to the traditional ball foot, the hook end effector furnished the stable shape needed to anchor the robot to the rungs and hang in tension with the center of mass outside of the support triangle, enabling it to traverse steeper ladders and withstand unmodeled perturbations to the base and feet. This finding emphasizes the importance of synergies between geometry and control policy for enhancing robot capabilities. 
Future work will focus on realizing quadrupeds that are capable of climbing up \textit{and down} ladders. Adding sensing modalities to the student training pipeline, such as depth camera images, will facilitate climbing ladders at industrial sites outside of the lab that are free standing, contain flat rungs, or have tapered designs.

\section{ACKNOWLEDGMENTS}

We thank Takahiro Miki and Fabian Tischhauser for their help. 
This work was supported by The Branco Weiss Fellowship - Society in Science, administered by ETH Zurich.


\end{document}